\documentclass[lettersize,journal]{IEEEtran}
\usepackage{amsmath,amsfonts}
\usepackage{algorithmic}
\usepackage{algorithm}
\usepackage{array}
\usepackage[caption=false,font=normalsize,labelfont=sf,textfont=sf]{subfig}
\usepackage{textcomp}
\usepackage{stfloats}
\usepackage{url}
\usepackage{verbatim}
\usepackage{graphicx}
\usepackage{cite}
\usepackage{multirow}
\usepackage{pifont}
\usepackage{eso-pic}
\newcommand\submittedtext{%
  \footnotesize This work has been submitted to the IEEE for possible publication. Copyright may be transferred without notice, after which this version may no longer be accessible.%
}

\begin{document}

\title{Calib3R: Hand-Eye Calibration and 3D Metric-Scaled Scene Reconstruction with 3D Foundation Models}

\author{Davide Allegro$^{1}$, Matteo Terreran$^{1}$ and Stefano Ghidoni$^{1}$
\thanks{$^{1}$All the authors are with the Department
of Information Engineering (DEI) at the University of Padova, via Gradenigo
6/B, 35131 Padova, Italy. 
        {\tt\small Email: [davide.allegro; matteo.terreran; stefano.ghidoni]@unipd.it}}%
}



\maketitle

\maketitle
\AddToShipoutPictureFG*{%
  \AtPageLowerLeft{%
    \makebox[\paperwidth]{%
      \hfill
      \raisebox{18pt}{
        \fbox{\parbox{\dimexpr0.55\paperwidth-2\fboxsep-2\fboxrule\relax}{\submittedtext}}%
      }%
      \hfill
    }%
  }%
}

\begin{abstract}
Robots often rely on RGB images for tasks like manipulation. However, reliable interaction typically requires a 3D scene representation that is metric-scaled and aligned with the robot reference frame. This depends on accurate hand-eye calibration and dense 3D reconstruction, tasks usually treated separately, despite both relying on geometric correspondences from RGB data. Traditional calibration techniques needs patterns, while RGB-based reconstruction yields 3D geometry with an unknown scale in an arbitrary frame. Multi-camera setups add further complexity, as data must be expressed in a shared reference frame.
We present Calib3R, a patternless method that jointly performs hand-eye calibration and metric-scaled 3D reconstruction via unified optimization. Calib3R handles single- and multi-camera setups on robot arms. It builds on a 3D foundation model to extract pointmaps from RGB images, which are combined with robot poses to reconstruct a scaled 3D scene aligned with the robot base reference frame. Experiments on diverse datasets show that Calib3R achieves accurate calibration with less than 10 images, outperforming patternless and pattern-based methods. Code will be available upon acceptance.
\end{abstract}


\section{Introduction}
Robotic systems rely on visual sensors to perceive and interact with their surroundings, with RGB and RGB-D cameras being among the most widely adopted sensing modalities for tasks such as manipulation~\cite{10874177,10418577,barcellona2025dream}. While RGB-D cameras provide direct depth measurements, their estimates can be noisy, incomplete, or unreliable under challenging real-world conditions~\cite{zhang2018deep}. RGB cameras, instead, are cheaper, widely available, and capable of capturing high-resolution appearance information. For this reason, several robotic applications increasingly rely on RGB observations alone~\cite{bauer2024challenges,an2024rgbmanip}.
However, RGB images do not directly provide the metric 3D information required for robotic manipulation. Therefore, reconstructing the workspace geometry from RGB observations becomes a fundamental step. The resulting 3D scene must then be metric-scaled and aligned with the robot base frame to enable effective interaction with the environment~\cite{chen2025vidbot}. 
This challenge is further amplified in multi-camera systems, where observations acquired from different viewpoints must additionally be expressed in a common robot coordinate frame to ensure geometric consistency~\cite{zhu2020autonomous,min2024multi}.

\begin{figure}[ht!]
  \centering
  \includegraphics[width=0.87\linewidth]{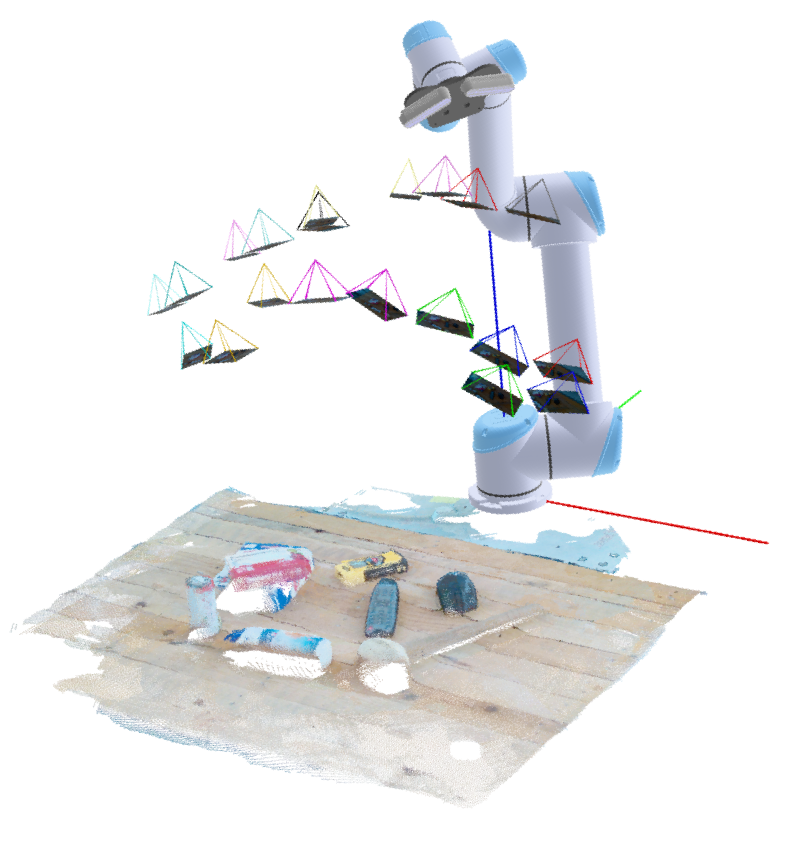}
  \vspace{-15pt}
  \caption{\textbf{Calib3R results}. Given RGB images and the corresponding robot poses, Calib3R combines local geometric predictions from a 3D foundation model with robot kinematic constraints to jointly estimate the camera-to-robot transformation, camera poses, and a dense metric 3D reconstruction. A single global optimization recovers all variables simultaneously, producing a reconstruction expressed in the robot reference frame.}
  \vspace{-12pt}
  \label{fig:calib3r_overview}
\end{figure}

Addressing this problem involves two tightly related tasks: (i) hand-eye calibration, which estimates the rigid camera-to-robot transformation, i.e., the pose of each camera with respect to the robot reference frame, and (ii) 3D scene reconstruction, which estimates the geometry of the observed environment. Although both tasks depend on geometric relationships across multiple RGB images, they are typically solved independently.

Hand-eye calibration methods commonly rely on fiducial markers or calibration patterns observed from multiple robot poses~\cite{5509954,9390394,allegro2024memroc,evangelista2023graph}. Other approaches require dedicated infrastructure, such as motion-capture systems~\cite{wang2022accurate}, or additional sensors, such as LiDAR~\cite{5509880,li2024automatic}. 
These requirements limit their applicability in realistic robotic environments, where calibration must be easily repeatable whenever the sensor configuration changes and should rely only on natural scene observations rather than dedicated hardware or markers.
Moreover, calibration methods designed for multi-camera systems often focus on estimating camera-to-camera extrinsics, while the alignment of each camera with the robot reference frame is either assumed known or addressed separately~\cite{rameau2022mc}.

At the same time, recent advances in data-driven 3D reconstruction enable recovering scene geometry directly from unconstrained RGB observations. In particular, 3D foundation models have demonstrated remarkable capabilities in estimating dense local geometry and camera motion from image collections. DUSt3R~\cite{wang2024dust3r} and MASt3R~\cite{leroy2024grounding} predict dense pointmaps and relative camera poses directly from image pairs, while MASt3R-SfM~\cite{duisterhof2024mast3r} extends this paradigm to multiple views through global alignment. Nevertheless, these reconstructions are not directly suitable for robotic applications, as they are defined up to an unknown scale and are not aligned with the robot reference frame. Robot poses naturally provide the metric and kinematic constraints needed to resolve both ambiguities, motivating a joint optimization of hand-eye calibration and dense 3D reconstruction.

Building on this observation, we propose Calib3R, a unified framework that jointly solves hand-eye calibration and dense metric 3D reconstruction from RGB images and corresponding robot poses. Given images acquired by a single- or multi-camera system mounted on a robot arm, Calib3R first leverages a 3D foundation model to estimate dense local pointmaps. These are then combined with robot motion constraints in a global optimization that jointly estimates the camera-to-robot transformation, camera poses, and a dense metric reconstruction expressed in the robot base frame. In the multi-camera setting, Calib3R additionally refines the inter-camera extrinsics within the same optimization, improving cross-view geometric consistency, as illustrated in Fig.~\ref{fig:calib3r_overview}.

Experiments on diverse real-world robotic setups show that Calib3R is accurate, robust, and data-efficient. It eliminates the need for depth sensors, fiducial markers, or multi-stage calibration procedures, while achieving high-precision calibration with fewer than ten images. Notably, Calib3R outperforms existing patternless methods and achieves accuracy comparable to, or better than, several pattern-based baselines.

In summary, the key contributions of this paper are:

\begin{itemize}
    \item Calib3R, a unified framework for joint hand-eye calibration and dense metric 3D reconstruction from RGB images and robot poses;
    \item a joint optimization that combines local geometric predictions from a 3D foundation model with robot kinematic constraints to estimate camera-to-robot transformations, camera poses, and dense 3D scene geometry without calibration patterns;
    \item a unified formulation for multi-camera robotic systems that jointly refines camera-to-robot and inter-camera extrinsics, producing a geometrically consistent reconstruction in the robot reference frame;
    \item an extensive evaluation on diverse real-world datasets demonstrating accuracy comparable to, or better than, pattern-based calibration methods while remaining robust even with few input images.
\end{itemize}
\vspace{-5pt}

\section{Related Works}
\label{sec:related_works}
\textbf{Hand-Eye Calibration.}
Hand-eye calibration estimates the rigid transformation between a camera and a robot reference frame typically through the classical AX=XB formulation~\cite{tsai1989new}. Closed-form and optimization-based solutions have been extensively studied~\cite{park1994robot,shah2013solving,li2010simultaneous}.
Most accurate methods rely on calibration targets and fiducial markers~\cite{evangelista2022unified,allegro2024multi}, or dedicated acquisition procedures~\cite{10186703}, while some multi-camera formulations require even external infrastructure such as motion-capture systems~\cite{wang2022accurate}. Other multi-camera calibration methods mainly estimate relative camera-to-camera extrinsics, while the alignment with the robot reference frame is often assumed known or solved separately~\cite{yan2022opencalib,furgale2013unified}. To reduce the dependence on calibration targets, SfM-based hand-eye calibration methods estimate camera motion from visual correspondences and exploit robot motion to recover the metric scale and the camera-to-robot transformation~\cite{andreff2001robot,heller2011structure}. However, these approaches estimate reconstruction and calibration independently, without jointly optimizing reconstruction and calibration parameters. Learning-based approaches have also been explored, but they often rely on specific assumptions, such as the visibility of the robot gripper in the camera view~\cite{valassakis2022learning}. Closest to our work, Zhi et al.~\cite{zhi2024unifying} proposed JCR, which leverages DUSt3R for patternless hand-eye calibration of a single camera. JCR follows a sequential pipeline, in which the visual reconstruction is first recovered independently of the robot and subsequently used as a fixed input for camera-to-robot alignment, with metric scale estimated through grid search. In contrast, Calib3R integrates robot kinematic constraints directly into the global reconstruction optimization, jointly refining camera poses, hand-eye transformations, and metric scale,  handling multiple onboard cameras through cross-camera rigidity constraints.

\textbf{3D Foundation Models.}
Traditional 3D reconstruction methods, such as Structure-from-Motion (SfM)~\cite{schonberger2016structure} and Multi-View Stereo (MVS)~\cite{furukawa2015multi}, rely on feature matching, camera pose estimation, triangulation, and multi-view optimization. Recent learning-based pipelines such as VGGSfM~\cite{wang2024vggsfm} and DF-SfM~\cite{he2024detector} improve this process, but still depend on classical geometric stages that can be fragile with few views or limited camera motion. Recent 3D foundation models introduce a different paradigm for RGB-based reconstruction. DUSt3R~\cite{wang2024dust3r} predicts dense pointmaps and relative camera poses from image pairs, subsequently aligned into a common coordinate system. MASt3R~\cite{leroy2024grounding} extends this formulation with dense pixel-level matching, while MASt3R-SfM~\cite{duisterhof2024mast3r} performs multi-view reconstruction through global alignment. VGGT~\cite{wang2025vggt} instead uses a feed-forward architecture to predict 3D scene geometry without explicit optimization. Despite their strong capabilities for 3D dense reconstruction, these models are not designed for robot-centric metric reconstruction. Their outputs are typically defined in an arbitrary reference frame and may lack reliable metric scale, limiting robotic applications that require accurate geometry aligned with the robot reference frame. Calib3R addresses these limitations by integrating robot poses into a unified robot-constrained optimization.



\section{Calib3R}
\label{sec:method}

\begin{figure*}[t]
  \centering
  \includegraphics[width=0.95\linewidth]{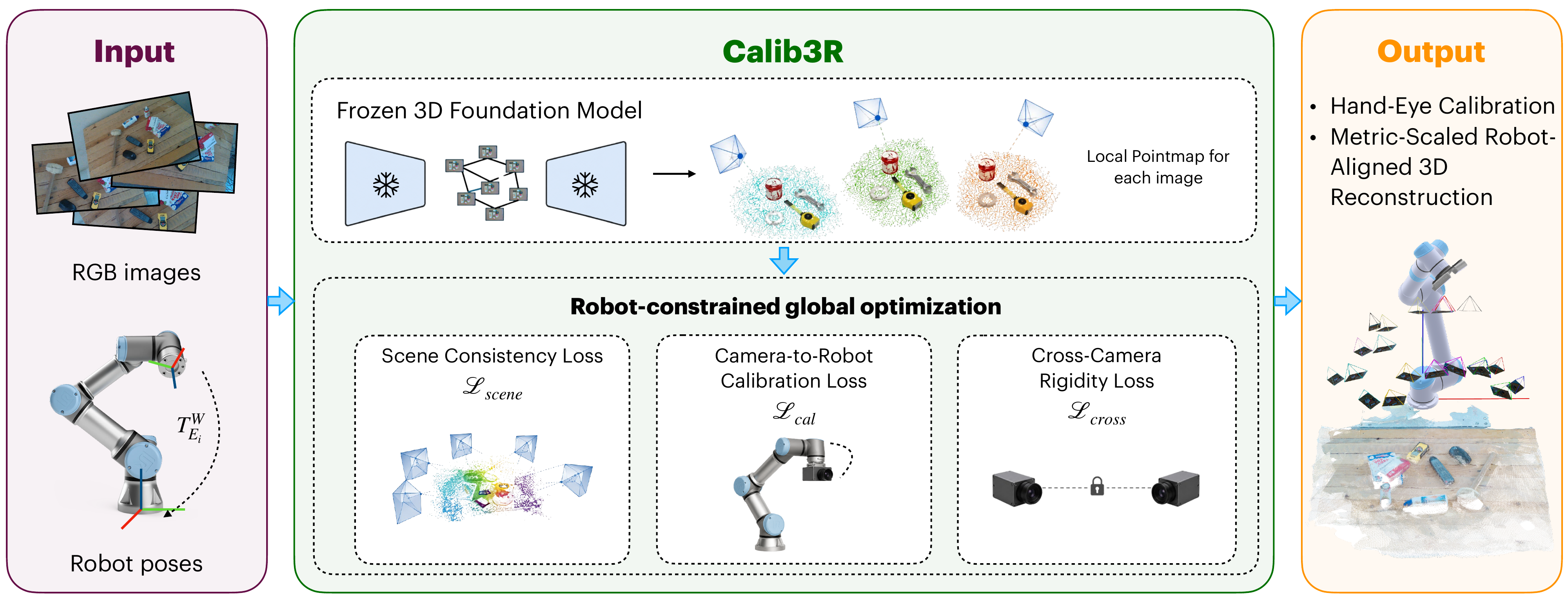}
  \vspace{-8pt}
  \caption{
  \textbf{Overview of the proposed Calib3R method.} 
  Given an $N \times M$ image set acquired by a robot equipped with $M$ cameras over $N$ poses, Calib3R constructs a sparse scene graph using frozen 3D foundation model to estimate local pointmap for each image. A global 3D optimization is then performed using a unified loss comprising: (i) a scene geometry term enforcing reconstruction consistency, (ii) a calibration term incorporating robot relative motions, and (iii) a cross-camera rigidity term. The optimization jointly estimates each camera’s trajectory, a per-camera scale factor, and the rigid camera-to-robot transformations, yielding a metric-scaled 3D reconstruction aligned with the robot reference frame.}
  \label{fig:calibe3r_pipeline}
  \vspace{-10pt}
\end{figure*}

Given a collection of RGB images acquired by one or multiple cameras mounted on a robot arm together with the corresponding robot poses, Calib3R jointly solves hand-eye calibration and dense metric 3D reconstruction within a unified global optimization. Unlike conventional pipelines that decouple visual reconstruction and hand-eye calibration, Calib3R incorporates robot kinematic constraints directly into the optimization, resolving the scale and reference-frame ambiguities of RGB-only reconstruction methods.

As illustrated in Fig.~\ref{fig:calibe3r_pipeline}, Calib3R integrates local geometric predictions from a frozen 3D foundation model within a robot-constrained global optimization. This optimization is driven by three complementary objectives: (i) a scene consistency loss, which aligns local geometry across views; (ii) a camera-to-robot calibration loss, which injects metric and kinematic constraints from the known robot motions; and (iii) a cross-camera rigidity loss, which enforces the fixed relative geometry between cameras mounted on the same robot. Jointly, these objectives recover a metric-scaled robot-aligned 3D reconstruction while simultaneously estimating the corresponding camera-to-robot transformations.
\vspace{-7pt}

\subsection{Local Geometric Predictions}
\label{sec:geometric_frontend}

Calib3R leverages a pretrained 3D foundation model to extract local geometric observations from RGB images. In the following, we describe the proposed optimization using the pointmap representation of MASt3R~\cite{leroy2024grounding}, which naturally provides dense local geometry, confidence estimates, and visual correspondences.
Specifically, given an image pair $\mathcal{I}^{u},\mathcal{I}^{v}\in\mathbb{R}^{W\times H\times 3}$, MASt3R predicts two dense pointmaps
$\mathbf{X}^{u,u},\mathbf{X}^{u,v}\in\mathbb{R}^{W\times H\times 3}$.
The pointmap $\mathbf{X}^{u,u}$ associates each pixel of $\mathcal{I}^{u}$ with a 3D point expressed in the camera frame of view $u$, while $\mathbf{X}^{u,v}$ associates pixels of $\mathcal{I}^{v}$ with 3D points expressed in the same frame. The model also outputs the corresponding confidence maps
$\mathbf{Q}^{u,u}$ and $\mathbf{Q}^{u,v}$,
alongside a set of dense pixel correspondences
$\mathcal{M}^{(u,v)}=\{y_{p}^{u} \leftrightarrow y_{p}^{v}\}_{p=1}^{|\mathcal{M}^{(u,v)}|}$.

For an image collection acquired by one or multiple cameras over multiple robot poses, each view is uniquely identified by the pair $u=(j,i)$, where $j\in\{0,\ldots,M-1\}$ denotes the camera and $i\in\{0,\ldots,N-1\}$ the robot pose. Using predicted co-visibility scores, we build a sparse overlap graph $\mathcal G=(\mathcal V,\mathcal E)$, where each vertex $u\in\mathcal V$ corresponds to an image $\mathcal{I}^{u}$ and an edge $(u,v) \in \mathcal{E}$ indicates sufficient visual overlap between $\mathcal{I}^u$ and $\mathcal{I}^v$.

Since an image may belong to multiple edges of the graph, multiple pointmap predictions can be available for it. For each image $\mathcal{I}^{u}$, these predictions are aggregated into a canonical pointmap $\tilde{\mathbf{X}}^{u}$ through confidence-weighted averaging:

\begin{equation}
\label{eq:canonical_pointmap}
\tilde{\mathbf{X}}^{u} =\frac{
\sum_{e\in\mathcal{E}^{u}}
\mathbf{Q}^{u,u}(e)\mathbf{X}^{u,u}(e)
}{
\sum_{e\in\mathcal{E}^{u}}
\mathbf{Q}^{u,u}(e)
},
\end{equation}
where $\mathcal{E}^{u}$ is the set of graph edges containing image $\mathcal{I}^{u}$.

The canonical pointmap $\tilde{\mathbf X}^u$ remains expressed in the local coordinate frame of view $u$. From it, we extract a depth map $\tilde Z^u$ and define the corresponding constrained pointmap through back-projection:
\begin{equation}
\label{eq:pointmap}
\boldsymbol{\chi}^{u}
=
\pi^{-1}
\left(
\sigma_{u},
K_{j},
T^{W}_{C_{j,i}},
\tilde{Z}^{u}
\right)\,.
\end{equation}
Here, $\sigma_{u}$ denotes the local scale variable associated with view $u$, $K_{j}$ denotes the intrinsic calibration matrix of the camera $C_{j}$, and $T^{W}_{C_{j,i}}$ transforms points from the local coordinate frame of the corresponding image $u$---acquired by camera $C_j$ at robot pose $i$---to the global reference frame $W$. Thus, $\tilde{\mathbf{X}}^{u}$ provides the fixed local geometric prediction, whereas $\boldsymbol{\chi}^{u}$ is its pose- and scale-dependent counterpart optimized by the robot-constrained global optimization.

Although MASt3R serves as the reference implementation throughout this section, Calib3R is not tied to any specific foundation model. The proposed optimization only assumes the availability of local geometric predictions and dense image correspondences, making it directly applicable to alternative 3D foundation models. Section~\ref{sec:backbone_multicam_ablation} validates this generality using MASt3R, DUSt3R, and VGGT.
\vspace{-4pt}

\subsection{Scene Consistency Loss}
\label{sec:scene_loss}

Since each constrained pointmap $\boldsymbol{\chi}^{u}$ defined in Sec.~\ref{sec:method} is expressed in the local coordinate frame of its corresponding view, the first step of the robot-constrained global optimization is to align all local reconstructions into a common reference frame. To ensure scalability, this alignment is performed only over the sparse co-visibility graph, whose edges connect image pairs with sufficient visual overlap. In multi-camera systems, the graph naturally includes both intra-camera and inter-camera pairs whenever different cameras observe common regions of the scene. The resulting scene consistency loss combines a 3D matching term and a 2D reprojection term:
\begin{equation}
\label{eq:scene_loss}
\mathcal{L}_{\mathrm{scene}}
=
\mathcal{L}_{3D}
+
\mathcal{L}_{2D}.
\end{equation}

The 3D matching term enforces consistency between corresponding 3D points across every connected pair of views:
\begin{equation}
\label{eq:master_3d_loss}
\mathcal{L}_{3D}
=
\sum_{\substack{
(u,v)\in\mathcal{E}\\
p\in\mathcal{M}^{(u,v)}
}}
q_p
\left\|
\boldsymbol{\chi}^{u}_{p}
-
\boldsymbol{\chi}^{v}_{p}
\right\|_2 ,
\end{equation}
where $q_p$ is the confidence associated with correspondence $p$. 

The 2D reprojection term further refines the alignment by enforcing that reconstructed 3D points project consistently onto their corresponding image observations:
\begin{equation}
\label{eq:master_2d_loss}
\begin{aligned}
\mathcal{L}_{2D}
=
\sum_{\substack{
(u,v)\in\mathcal{E}\\
(y_p^u,y_p^v)\in\mathcal{M}^{(u,v)}
}}
&
q_p
\bigg(
\left\|
y_p^u-\pi_u\!\left(\boldsymbol{\chi}^{v}_{p}\right)
\right\|_2
\\
&+
\left\|
y_p^v-\pi_v\!\left(\boldsymbol{\chi}^{u}_{p}\right)
\right\|_2
\bigg)\,.
\end{aligned}
\end{equation}

Together, these visual constraints jointly optimize camera poses and scene geometry across all connected views. However, since they rely solely on visual information, the recovered reconstruction remains defined up to an unknown scale and an arbitrary global reference frame.
\vspace{-5pt}

\subsection{Camera-to-Robot Calibration Loss}
\label{sec:calibration_loss}

\begin{figure*}[t]
  \centering
  \includegraphics[width=0.8\linewidth]{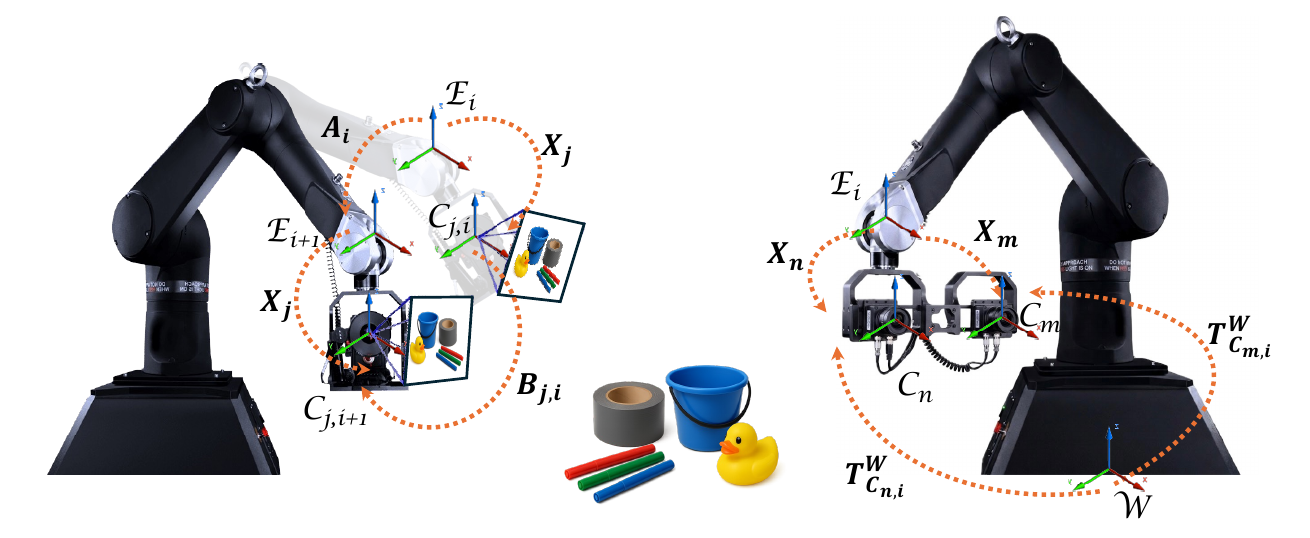}
  \vspace{-10pt}
  \caption{\textbf{Formulation of the hand-eye calibration problem.} (Left) Classical single-camera formulation. Let $E_i$ denote the robot end-effector pose at time step $i$. The robot end-effector moves from pose $i$ to $i+1$, producing the relative robot motion $A_i$, while the camera acquires images of a static scene, yielding the corresponding relative camera motion $B_{j,i}$. The unknown transformation $X_{j}$ represents the camera-to-robot transformation. (Right) Multi-camera formulation adopted in Calib3R. Multiple rigidly attached cameras $C_n$ and $C_m$ simultaneously observe the scene from synchronized viewpoints. Their absolute poses $T^{W}_{C_{n,i}}$ and $T^{W}_{C_{m,i}}$ are jointly optimized under the shared robot kinematic constraints and the rigid camera configuration.}
  \label{fig:hec_scheme}
  \vspace{-10pt}
\end{figure*}

To recover metric scale and align the reconstruction with the robot frame, Calib3R introduces a calibration loss inspired by the classical hand-eye formulation, as illustrated in Fig.~\ref{fig:hec_scheme}. 
In particular, given two consecutive robot poses, the relative motion of the robot end-effector $E$ is:
\begin{equation}
\label{eq:robot_motion}
A_i=
T^{{E}_i}_{{E}_{i+1}}
=
\left(T^{{W}}_{{E}_i}\right)^{-1}
T^{{W}}_{{E}_{i+1}},
\quad i=0,\dots,N-1 \,.
\end{equation}
Similarly, the visual relative motion of camera $C_j$ is computed from the camera poses variables:
\begin{equation}
\label{eq:camera_motion}
B_{j,i}=
T^{{C}_{j,i}}_{{C}_{j,i+1}}=
\left(T^{{W}}_{{C}_{j,i}}\right)^{-1}
T^{{W}}_{{C}_{j,i+1}} \,.
\end{equation}

In standard hand-eye calibration, the camera relative motion $B_{j,i}$ is metric because it is usually estimated from observations of a known calibration target. In contrast, in our setting the camera motion is recovered from RGB-only reconstruction and is therefore defined only up to an unknown scale. We address this by introducing an optimization variable $\lambda_j>0$ for each camera and applying it to the translational component of the visual relative motion:
\begin{equation}
\label{eq:scaled_camera_motion}
B_{j,i}(\lambda_j)=
\begin{bmatrix}
R_{B_{j,i}} & \lambda_j t_{B_{j,i}} \\
\mathbf{0}^{\top} & 1
\end{bmatrix}\,.
\end{equation}

Therefore, the calibration loss for camera $j$ can be defined as:
\begin{equation}
\label{eq:revised_hec}
\mathcal{L}_{\mathrm{cal},j}
=
\sum_{i=0}^{N-1}
\left\|
A_iX_j -X_jB_{j,i}(\lambda_j) 
\right\|_2 \,,
\end{equation}
where $X_j=T^{E}_{C_{j}}$ is the unknown camera-to-robot transformation that forces the visual camera motion to be consistent with the robot motion
Consequently, the hand-eye calibration and the metric scale are recovered during reconstruction together with camera poses, rather than being estimated in a separate post-processing step.
\vspace{-5pt}

\subsection{Cross-Camera Rigidity Loss}
\label{sec:cross_camera_loss}

The scene consistency loss already exploits visual overlap between different cameras whenever inter-camera image pairs are connected in the co-visibility graph. However, cameras rigidly mounted on the same robotic system satisfy an additional physical constraint: their relative pose remains constant throughout the robot motion, independently of the observed scene. Calib3R exploits this property through a cross-camera rigidity loss. Unlike the robot-constrained calibration loss, which aligns the \emph{temporal} motion of each individual camera with the corresponding robot motion, the proposed cross-camera loss enforces \emph{spatial} consistency between different cameras at each synchronized acquisition time.

Let $X_n=T^{E}_{C_n}$ and $X_m=T^{E}_{C_m}$ be the camera-to-robot transformations of cameras $C_n$ and $C_m$, respectively. Their relative transformation induced by the estimated hand-eye parameters is
$X_{n,m}=X_n^{-1}X_m$.

At each acquisition time $i$, the camera motion can be defined as:
\begin{equation}
T^{C_{n,i}}_{C_{m,i}}
=
\left(T^{W}_{C_{n,i}}\right)^{-1}
T^{W}_{C_{m,i}}\,,
\end{equation}

whose translation is converted into metric units using the estimated scale factors of the corresponding cameras.

The cross-camera rigidity loss enforces consistency between the relative pose recovered from the visual reconstruction and that induced by the estimated camera-to-robot transformations. Specifically, we optimize

\begin{equation}
\label{eq:joint_loss}
\mathcal{L}_{\mathrm{cross}}
=
\sum_{n<m}
\sum_{i=0}^{N-1}
\left|\left|
X_{n,m}- 
T^{C_{n,i}}_{C_{m,i}}
\right|\right|_{2}\,.
\end{equation}

This loss couples all cameras through their common rigid mounting while remaining fully compatible with the visual reconstruction objective. Consequently, it provides complementary geometric constraints regardless of the amount of visual overlap between cameras, improving the overall consistency of the estimated calibration and reconstruction.
\vspace{-5pt}

\subsection{Global Optimization}
\label{sec:global_optimization}




The final Calib3R objective combines scene consistency, camera-to-robot calibration, and cross-camera rigidity:
\begin{equation}
\label{eq:master_hec_joint_loss}
\mathcal{L}_{\mathrm{Calib3R}}=
\mathcal{L}_{\mathrm{scene}}
+
\sum_{j=0}^{M-1}
\mathcal{L}_{\mathrm{cal},j}
+
\alpha_{\mathrm{cross}}
\mathcal{L}_{\mathrm{cross}},
\end{equation}
where $\alpha_{\mathrm{cross}}$ is used to enable cross-camera rigidity constraint.

The objective is jointly optimized with respect to the camera poses $T^{W}_{C_{j,i}}$, camera-to-robot transformations $T^{E}_{C_{j}}$ and per-camera metric scale factors $\lambda_{j}$. For single-camera systems, the same formulation applies by setting $\alpha_{\mathrm{cross}}=0$. The optimization directly recovers a metric-scaled 3D reconstruction with corresponding camera trajectories aligned with the robot reference frame and camera-to-robot transformations within a unified robot-centric framework.

\section{Experimental Evaluation Protocol}
\begin{figure*}[t]
  \centering
  \begin{minipage}[b]{0.18\linewidth}
    \includegraphics[width=\linewidth]{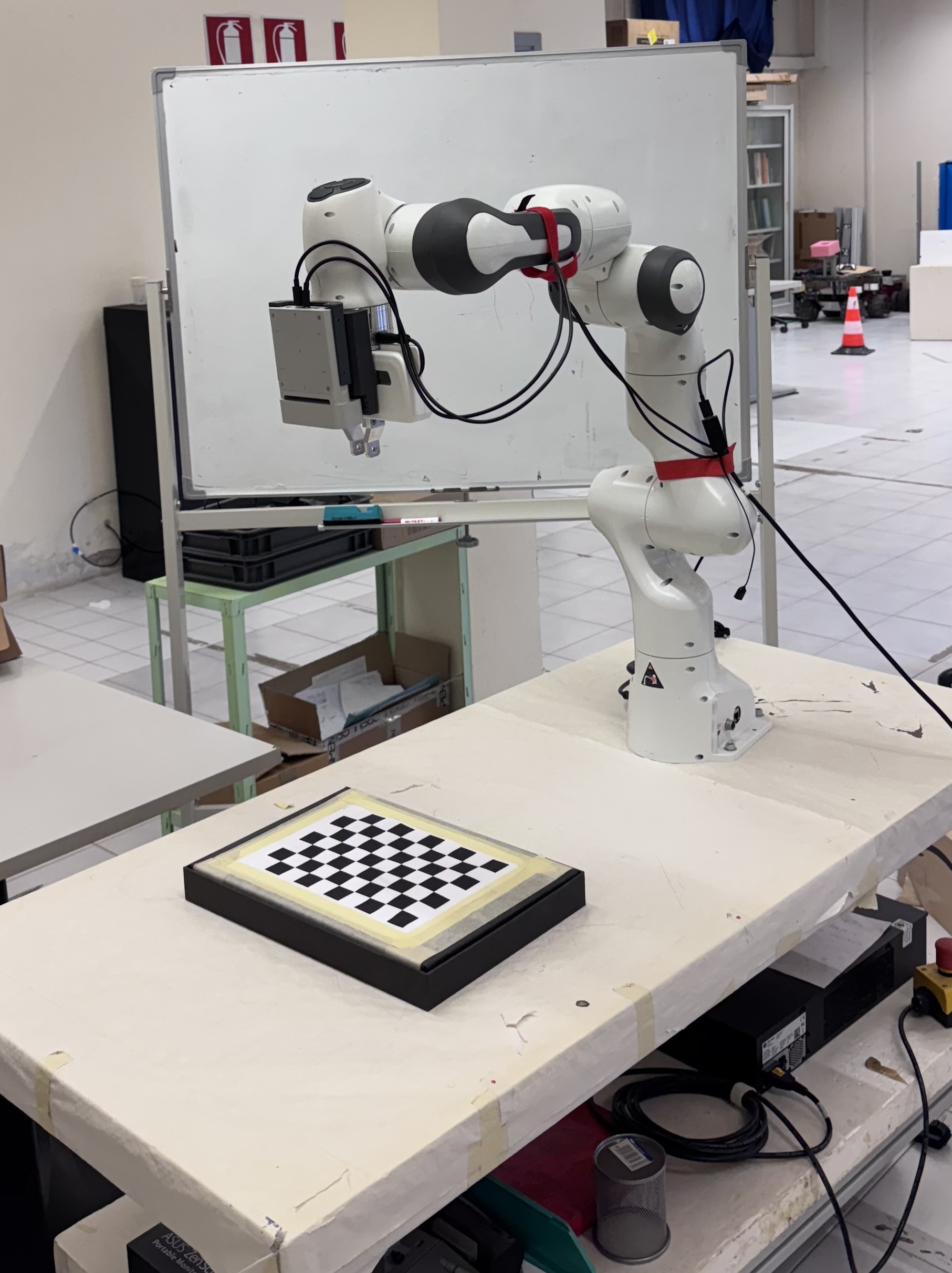}
    \caption{Franka Pattern dataset: experimental setup with a 7×10 checkerboard (3\,cm squares) placed within the robot’s workspace for calibration.}
    \label{fig:franka_dataset_A}
  \end{minipage}
  \hfill
  \begin{minipage}[b]{0.182\linewidth}
    \includegraphics[width=\linewidth]{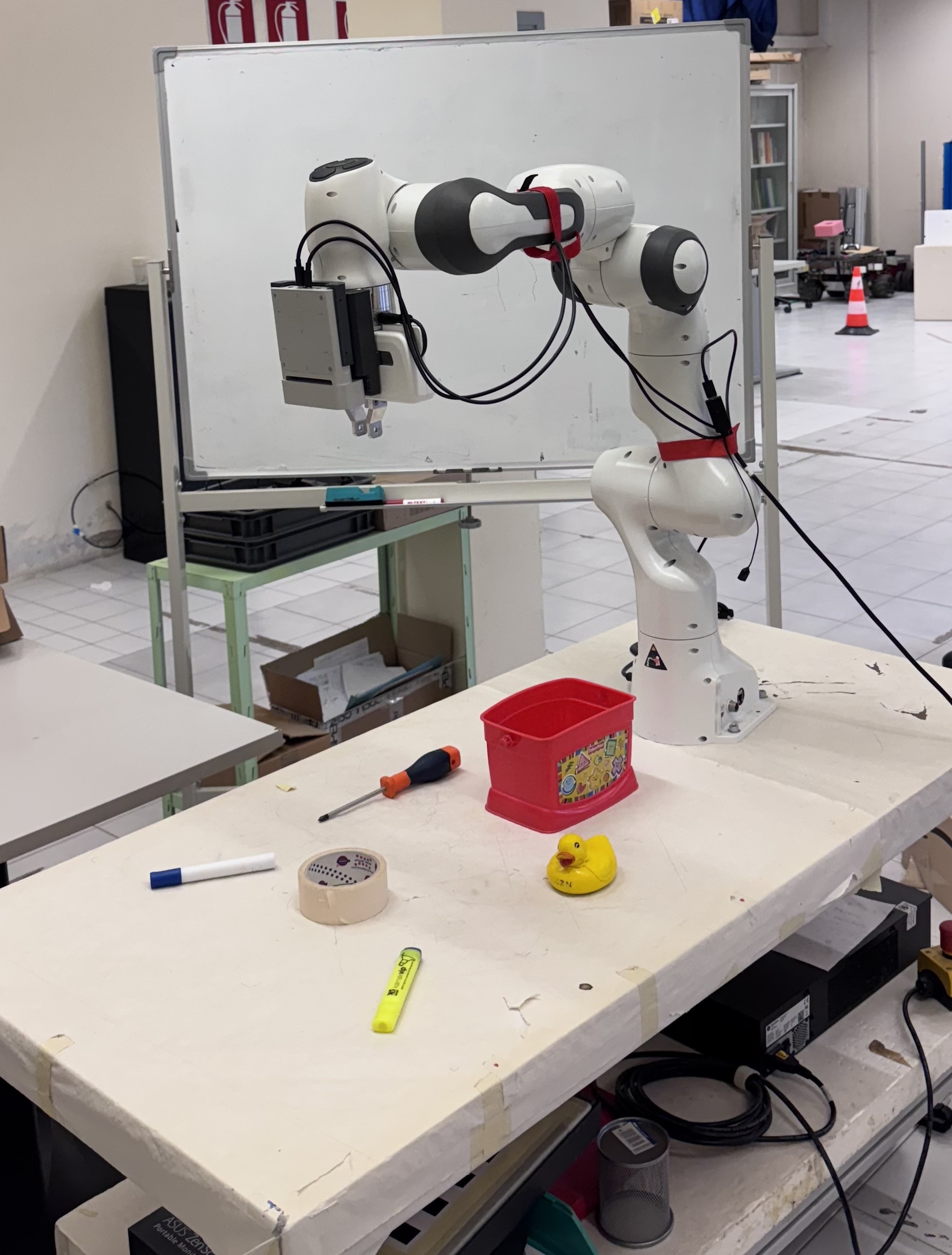}
    \caption{Franka Object dataset: generic objects are placed in the robot’s workspace to support both calibration and scene reconstruction.}
    \label{fig:franka_dataset_B}
  \end{minipage}
  \hfill
  \begin{minipage}[b]{0.18\linewidth}
    \includegraphics[width=\linewidth]{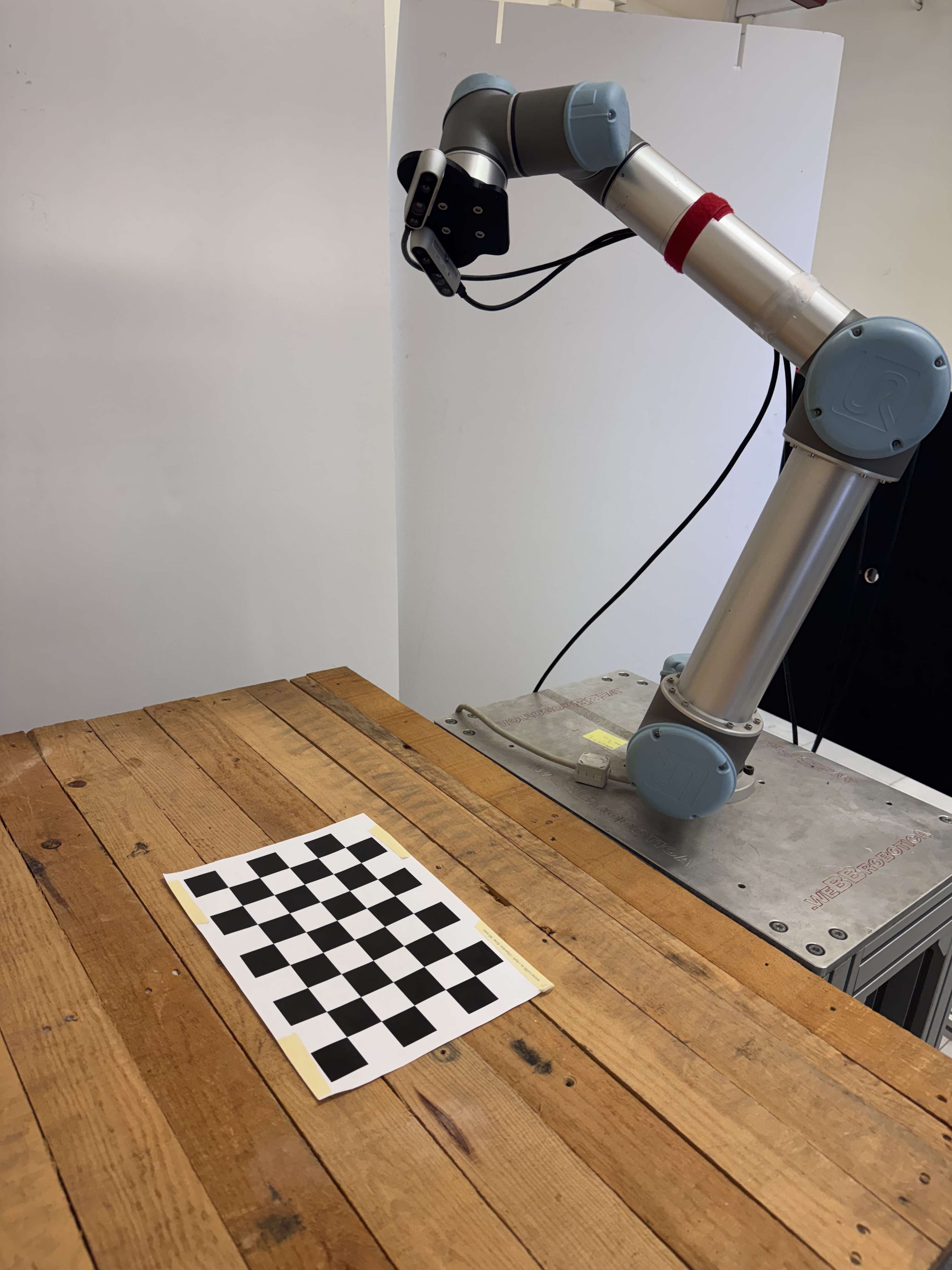}
    \caption{UR Pattern dataset: experimental setup with a 6×9 checkerboard (3\,cm squares) placed within the robot’s workspace for multi-camera calibration.}
    \label{fig:ur_dataset_A}
  \end{minipage}
  \hfill
  \begin{minipage}[b]{0.18\linewidth}
    \includegraphics[width=\linewidth]{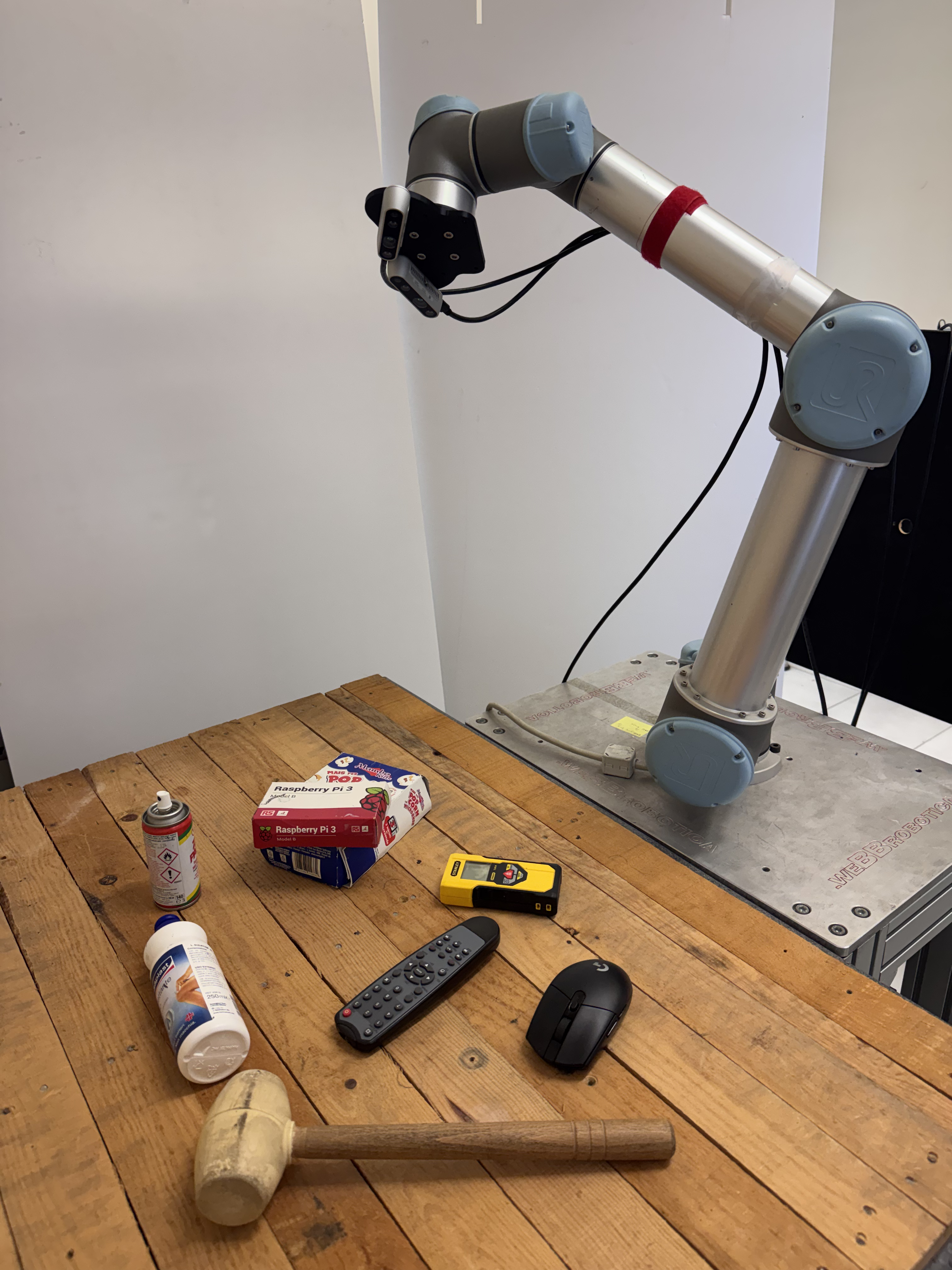}
    \caption{UR Object dataset: generic objects are placed in the robot’s workspace to support both multi-camera calibration and scene reconstruction.}
    \label{fig:ur_dataset_B}
  \end{minipage}
  \hfill
  \begin{minipage}[b]{0.178\linewidth}
    \includegraphics[width=\linewidth]{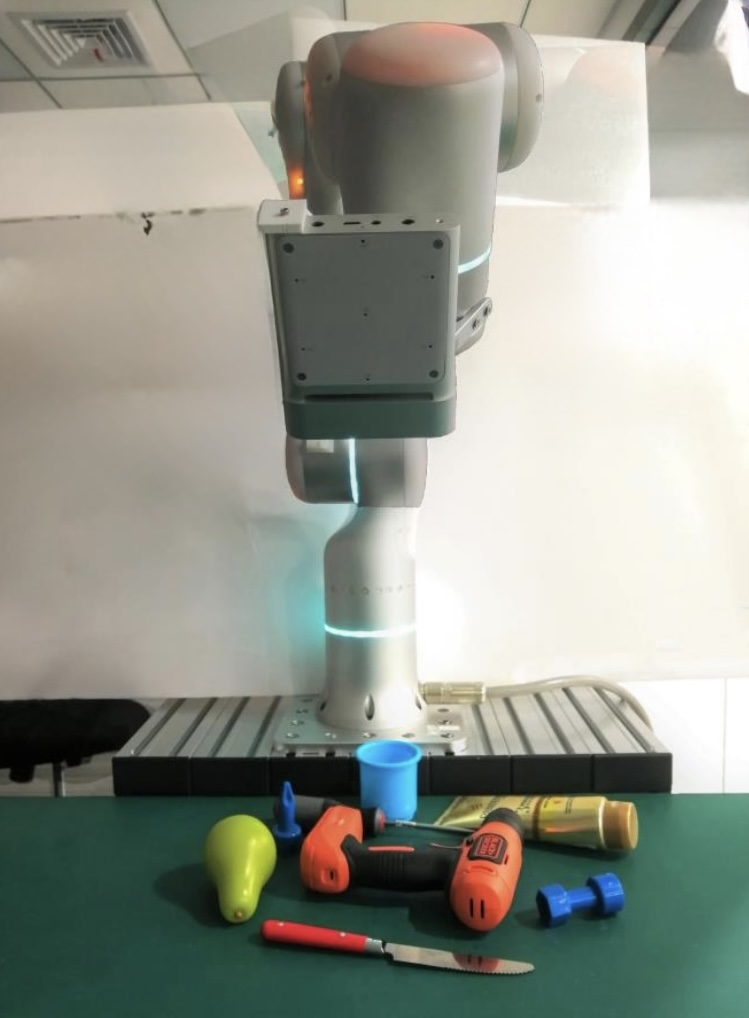}
    \caption{GraspNet-1Billion dataset: robot setup with generic objects placed in its workspace~\cite{fang2020graspnet}.}
    \label{fig:graspnet_dataset}
  \end{minipage}
  \vspace{-7pt}
\end{figure*}


We evaluate Calib3R on both calibration-pattern and patternless robotic datasets. The former support quantitative comparison with pattern-based calibration methods and provide a known metric reference, while the latter assess performance in realistic manipulation scenarios. We evaluate both hand-eye calibration accuracy and the metric accuracy of the reconstructed scene.

\subsection{Datasets}
We evaluate Calib3R on five robotic-arm datasets, including four datasets acquired using our own setups and one public benchmark. 
The \textit{Franka Pattern} and \textit{Franka Object} datasets were acquired with a Kinect Azure mounted on a Franka Emika Panda robot (Fig.~\ref{fig:franka_dataset_A}-\ref{fig:franka_dataset_B}). Both contain 25 robot poses; the former uses a \(7\times10\) checkerboard with 3\,cm squares, while the latter replaces it with generic tabletop objects. Similarly, the \textit{UR Pattern} and \textit{UR Object} datasets were acquired with two Intel RealSense D435 cameras mounted on the end-effector of a UR5 (Fig.~\ref{fig:ur_dataset_A}-\ref{fig:ur_dataset_B}). They contain 15 robot poses using a \(6\times9\) checkerboard with 3\,cm squares and generic objects, respectively. We additionally evaluate Calib3R on the public GraspNet-1Billion benchmark~\cite{fang2020graspnet} (Fig.~\ref{fig:graspnet_dataset}) using 25 images from scene\_0100.

\subsection{Metrics}
We evaluate two complementary aspects of the proposed method: hand-eye calibration accuracy and metric reconstruction accuracy.

\textbf{Calibration accuracy} is evaluated using the translation error $e_t$ and the rotation error $e_\theta$ between the estimated hand-eye transformation $(\hat{R},\hat{t})$ and the corresponding ground truth $(R,t)$:

\begin{equation}
\label{eq:metric_ev}
e_{t} = \frac{1}{M}\sum_{i=0}^{M-1}\|t - \hat{t}\|_{2}, \quad e_{\theta} = \frac{1}{M}\sum_{i=0}^{M-1}angle(R^T \hat{R})\,,
\end{equation}

where $angle(\cdot)$ denotes the angle derived from the axis-angle representation of the relative rotation matrix $R^T \hat{R}$.

\textbf{Metric scale accuracy} is evaluated on the 
datasets containing a checkerboard, namely Franka Pattern and UR Pattern. Let $d^{(k)}$ denote the mean distance between adjacent reconstructed checkerboard corners for detection $k$. We compute the average reconstructed square size $m_s$ and report the absolute scale error $m_s-s$, the standard deviation $\sigma_s$, and the relative scale error
\begin{equation}
m_s=\frac{1}{K}\sum_{k=0}^{K-1}d^{(k)},\qquad
\delta_s(\%)=\frac{|m_s-s|}{s}\times100,
\end{equation}
where $s$ denotes the ground-truth checkerboard square size.

\subsection{Computational Setup}
Calib3R and the compared foundation model-based approaches are executed on an NVIDIA A40 GPU (48\,GB), while classical calibration baselines are run on an Intel Core i7-1280P CPU with 16\,GB RAM.

\section{Experimental Results}

\begin{table*}[ht!]
\caption{Average error of the estimated camera-to-robot transformation.
The Pattern column indicates whether the method requires a calibration target.
Translation errors are reported in centimetres and rotation errors in radians.}
\label{tab:h2e_results}
\centering
\resizebox{\textwidth}{!}{%
\begin{tabular}{l|c|cc|cc|cc|cc|cc}
\hline
\hline
\multirow{2}{*}{Method} & \multirow{2}{*}{Pattern} 
& \multicolumn{2}{c|}{Franka Pattern} 
& \multicolumn{2}{c|}{Franka Object} 
& \multicolumn{2}{c|}{UR Pattern} 
& \multicolumn{2}{c|}{UR Object} 
& \multicolumn{2}{c}{GraspNet-1Billion~\cite{fang2020graspnet}} \\
& 
& $e_{t}$ [cm] & $e_{\theta}$ [rad] 
& $e_{t}$ [cm] & $e_{\theta}$ [rad] 
& $e_{t}$ [cm] & $e_{\theta}$ [rad] 
& $e_{t}$ [cm] & $e_{\theta}$ [rad] 
& $e_{t}$ [cm] & $e_{\theta}$ [rad] \\
\hline

Tsai~\cite{tsai1989new} & \ding{51} 
& 2.631 & 0.041 
& - & - 
& 1.672 & 0.021
& - & - 
& - & - \\

Park~\cite{park1994robot} & \ding{51} 
& 3.610 & 0.038 
& - & - 
& 2.113 & 0.021 
& - & - 
& - & - \\

Horaud~\cite{horaud1995hand} & \ding{51} 
& 2.743 & 0.038 
& - & - 
& 2.882 & 0.031
& - & - 
& - & - \\

Andreff~\cite{andreff1999line} & \ding{51} 
& 11.341 & 0.033 
& - & - 
& 3.912 & 0.028 
& - & - 
& - & - \\

Daniilidis~\cite{daniilidis1996dual} & \ding{51} 
& 2.472 & 0.035 
& - & - 
& 4.311 & 0.028
& - & - 
& - & - \\

Evangelista~\cite{evangelista2022unified} & \ding{51} 
& \pmb{0.781} & 0.041 
& - & - 
& \pmb{0.691} & \underline{0.015} 
& - & - 
& - & - \\

\hline
\hline

JCR w/ COLMAP~\cite{schonberger2016structure} & \ding{55} 
& 28.210 & 0.122 
& 18.212 & 0.092 
& 35.123 & 0.106
& 17.123 & 0.102
& 22.324 & 0.122 \\

JCR w/ DUSt3R~\cite{wang2024dust3r} & \ding{55} 
& 24.641 & 0.043 
& 16.431 & 0.031 
& 11.123 & 0.022
& 2.912 & 0.031
& 18.212 & 0.042 \\

JCR w/ MASt3R SfM~\cite{duisterhof2024mast3r} & \ding{55} 
& 1.812 & \underline{0.023} 
& 1.742 & \underline{0.027} 
& 4.471 & 0.025
& 1.323 & 0.027
& 3.427 & 0.032 \\

JCR w/ VGGT~\cite{wang2025vggt} & \ding{55} 
& 1.734 & 0.034 
& \underline{1.674} & 0.038 
& 4.123 & 0.023
& \underline{1.231} & \underline{0.018}
& \underline{2.442} & \underline{0.031} \\

Calib3R & \ding{55} 
& \underline{1.127} & \pmb{0.014} 
& \pmb{0.415} & \pmb{0.011} 
& \underline{1.512} & \pmb{0.014}
& \pmb{0.636} & \pmb{0.017}
& \pmb{1.744} & \pmb{0.023} \\

\hline
\hline
\end{tabular}%
}
\vspace{-10pt}
\end{table*}

The experimental evaluation examines the calibration and reconstruction performance of Calib3R, its robustness to a limited number of input views, and the impact of the geometric backbone and multi-camera formulation.

\subsection{Hand-Eye Calibration Accuracy}
\label{sec:calibration_results}


We compare Calib3R with both classical pattern-based calibration methods and recent patternless approaches. Classical baselines include Tsai~\cite{tsai1989new}, Park~\cite{park1994robot}, Horaud~\cite{horaud1995hand}, Andreff~\cite{andreff1999line}, Daniilidis~\cite{daniilidis1996dual}, and Evangelista~\cite{evangelista2022unified}. For the patternless baselines, camera trajectories are first reconstructed using COLMAP~\cite{schonberger2016structure}, DUSt3R~\cite{wang2024dust3r}, MASt3R-SfM~\cite{duisterhof2024mast3r}, or VGGT~\cite{wang2025vggt}, and then provided to JCR~\cite{zhi2024unifying} for hand-eye calibration. These two-stage pipelines are compared with our proposed unified optimization. Unless otherwise specified, Calib3R uses MASt3R as its geometric backbone, while the UR datasets are evaluated using the complete multi-camera formulation with the proposed cross-camera rigidity loss.

Table~\ref{tab:h2e_results} reports the average translation and rotation errors of the estimated camera-to-robot transformations. We first consider the Franka Pattern and UR Pattern datasets, where dedicated checkerboards allow direct comparison with classical calibration methods. Calib3R achieves the lowest rotation error in both datasets while remaining competitive in translation. Compared with reconstruction-based approaches, Calib3R consistently outperforms all two-stage baselines. The comparison with JCR using MASt3R-SfM is particularly informative, since both methods rely on the same underlying 3D foundation model. The consistent improvement in both translation and rotation accuracy demonstrates that jointly optimizing reconstruction and calibration is more effective than estimating them sequentially. Calib3R also outperforms JCR combined with VGGT, reducing the translation error by at least 28\%, even though the competing pipeline employs a more recent visual backbone. Among classical calibration methods, only Evangelista et al.~\cite{evangelista2022unified} achieve slightly lower translation errors, whereas Calib3R obtains the lowest rotation error on both datasets, reaching only 0.014\,rad without requiring a dedicated calibration procedure.

We next evaluate Calib3R on the corresponding Franka Object and UR Object datasets, where the checkerboard is replaced by generic tabletop objects while preserving exactly the same robot trajectories. In this setting, classical pattern-based methods are no longer applicable. Calib3R achieves the lowest translation and rotation errors among all patternless methods and, remarkably, its translation errors are also lower than the best pattern-based results obtained on the corresponding pattern-based dataset for each robot setup. This behaviour is consistent with the characteristics of modern 3D foundation models, which benefit from the richer textures and more diverse geometric structures provided by natural objects than by repetitive planar calibration patterns. Consequently, the visual backbone establishes more reliable dense correspondences, supporting accurate calibration directly from natural scene observations without requiring dedicated calibration patterns.

Finally, we evaluate Calib3R on the public GraspNet-1Billion benchmark~\cite{fang2020graspnet}. The results confirm the same trend, with Calib3R achieving the best calibration accuracy among all applicable patternless methods. The slightly higher errors compared with the Franka Object dataset are consistent with the more limited motion diversity of the selected GraspNet sequence, whose viewpoints are predominantly top-down and exhibit only limited roll and pitch variations, reducing the observability of the hand-eye parameters.
\vspace{-5pt}

\subsection{Robustness to the Number of Views}
\label{sec:number_views}


We further evaluate the robustness of Calib3R to the number of available views on the Franka Object dataset. Starting from the complete sequence of 25 images, we progressively remove two randomly selected views at each step until only five remain. We compare Calib3R against the patternless two-stage baselines using DUSt3R, MASt3R-SfM, and VGGT.
\begin{figure}[t!]
  \centering
  \includegraphics[width=1.0\linewidth]{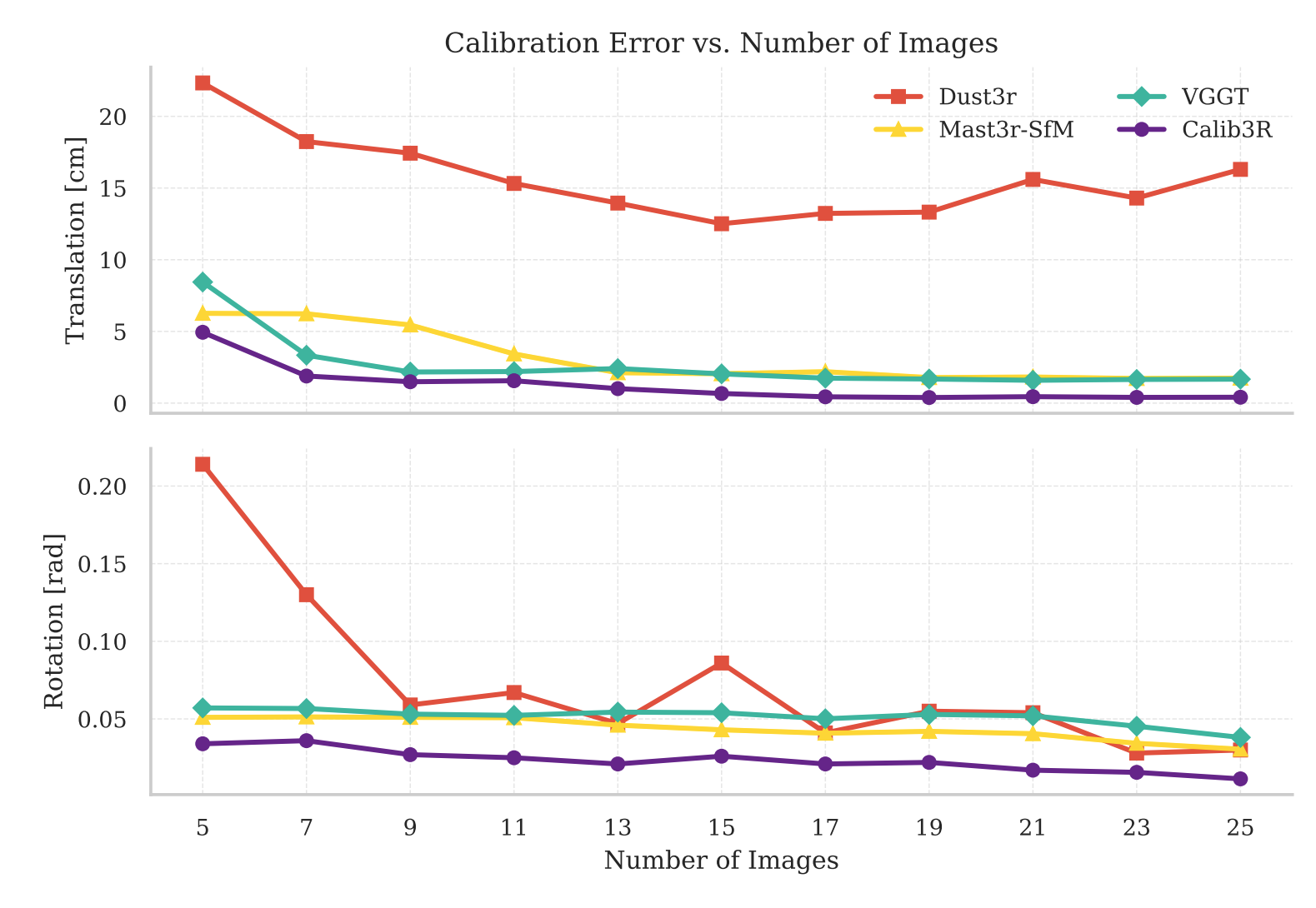}
  \vspace{-20pt}
  \caption{Camera-to-robot transformation accuracy as a function of the number of input views on the Franka Object dataset. Starting from 25 images, two randomly selected views are removed at each step until only five remain.}
  \label{fig:h2e_calib_vs_images}
\end{figure}

Figure~\ref{fig:h2e_calib_vs_images} shows that Calib3R achieves the lowest translation and rotation errors across the entire range of input views. Even with five images, the translation error remains below 5\,cm and the rotation error below 0.04\,rad. With 15 images, the translation error falls below 1\,cm. These results indicate that the proposed joint optimization effectively exploits the available geometric and robot-motion constraints, remaining accurate even when a small number of views is available.
\vspace{-5pt}

\subsection{Metric Reconstruction Accuracy}
\label{sec:metric_scale_results}

We evaluate the metric accuracy of the reconstructed scene on the Franka Pattern dataset. Since this dataset contains a checkerboard with known dimensions, it enables a direct evaluation of the recovered metric scale. We compare Calib3R with JCR combined with DUSt3R, MASt3R-SfM, and VGGT. COLMAP is excluded because its sparse reconstruction does not provide sufficiently dense geometry to reliably measure checkerboard square dimensions in 3D.

\begin{table}[t]
\caption{Metric reconstruction accuracy on the Franka Pattern dataset.
We report the mean scale error, its standard deviation, and the relative
error with respect to the true checkerboard square size of 3\,cm.
Lower values are better.}
\label{tab:metric_scale_accuracy}
\centering
\begin{tabular}{l|c|c|c}
\hline
\hline
\multirow{2}{*}{Method} & \multicolumn{3}{c}{Franka Pattern dataset} \\
& $m_s-s$ [cm] & $\sigma_s$ [cm] & $\delta_s$ [\%] \\
\hline
JCR w/ DUSt3R~\cite{wang2024dust3r}
& 0.79 & 0.091 & 26.33 \\
JCR w/ MASt3R-SfM~\cite{duisterhof2024mast3r}
& 0.41 & 0.027 & 13.67 \\
JCR w/ VGGT~\cite{wang2025vggt}
& \underline{0.17} & \underline{0.008} & \underline{5.67} \\
Calib3R
& \pmb{0.11} & \pmb{0.005} & \pmb{3.67} \\
\hline
\hline
\end{tabular}
\vspace{-10pt}
\end{table}

As reported in Table~\ref{tab:metric_scale_accuracy}, Calib3R achieves the highest metric accuracy, reducing the relative scale error from 5.67\% for the strongest baseline, JCR with VGGT, to 3.67\%. Moreover, the lower standard deviation indicates a more consistent metric reconstruction across the scene. These results show that incorporating robot kinematic constraints into the proposed joint optimization improves not only camera-to-robot calibration, but also the metric accuracy and consistency of the reconstructed geometry.

\subsection{Geometry Backbone and Multi-Camera Ablation}
\label{sec:backbone_multicam_ablation}


We finally evaluate two complementary aspects of Calib3R: its dependence on the underlying 3D foundation model and the contribution of the proposed cross-camera rigidity loss. Besides MASt3R, we integrate DUSt3R and VGGT by converting their geometric predictions into a common representation of local pointmaps and dense pairwise correspondences required by the optimizer. For DUSt3R, we directly use the predicted pairwise pointmaps and extract correspondences weighted by the associated confidence maps. For VGGT, which predicts a globally consistent reconstruction, we transform the predicted per-view geometry into the corresponding local camera frames and derive dense pairwise correspondences using the confidence scores. This ensures that all backbones provide the same local geometric representation and can therefore be evaluated within the same Calib3R optimization.

\begin{table}[t]
\caption{Ablation of the geometry backbone and multi-camera rigidity term.
We report the mean camera-to-robot error over the onboard cameras.}
\label{tab:calib3r_backbone_multicamera_ablation}
\centering
\setlength{\tabcolsep}{4pt}
\renewcommand{\arraystretch}{0.95}
\begin{tabular}{l|cc|cc|cc}
\hline
\hline
\multirow{3}{*}{Setting} &
\multicolumn{6}{c}{UR Object dataset} \\
\cline{2-7}
& \multicolumn{2}{c|}{\rule{0pt}{2.6ex}DUSt3R} &
\multicolumn{2}{c|}{\rule{0pt}{2.6ex}MASt3R} &
\multicolumn{2}{c}{\rule{0pt}{2.6ex}VGGT} \\[2pt]
& $e_t$ [cm] & $e_\theta$ [rad] &
  $e_t$ [cm] & $e_\theta$ [rad] &
  $e_t$ [cm] & $e_\theta$ [rad] \\[2pt]
\hline
w/o MC & 1.181 & 0.018 & 0.792 & 0.017 & 0.574 & 0.014 \\
w/ MC  & \textbf{1.082} & \textbf{0.015} &
          \textbf{0.636} & \textbf{0.014} &
          \textbf{0.403} & \textbf{0.014} \\
\hline
\hline
\end{tabular}
\vspace{-10pt}
\end{table}


We perform this analysis on the UR Object dataset, exploiting its multi-camera configuration to quantify the contribution of the proposed cross-camera rigidity loss across different 3D foundation models. Table~\ref{tab:calib3r_backbone_multicamera_ablation} shows that Calib3R produces accurate calibration estimates with all three backbones, confirming that the proposed optimization is not tied to a specific 3D foundation model. Although the formulation is presented using MASt3R, whose pairwise pointmap representation naturally motivated the proposed optimization, the objective only requires local geometry and dense visual correspondences. Consequently, newer foundation models can be seamlessly integrated as they become available. In particular, VGGT achieves the highest calibration accuracy, demonstrating that Calib3R can directly benefit from advances in 3D foundation models without modifying the optimization itself.

Finally, enabling the proposed cross-camera rigidity loss consistently improves translation accuracy for all evaluated backbones while preserving or improving rotation accuracy. This confirms that explicitly enforcing the fixed relative geometry between onboard cameras provides complementary constraints independently of the underlying geometric backbone.






\section{Conclusions}

We presented Calib3R, a unified framework for camera-to-robot calibration and metric 3D reconstruction from RGB images and robot poses. By incorporating robot kinematic constraints into pointmap-based optimization, Calib3R resolves visual reconstruction, metric scale, and camera-to-robot alignment in a common robot-centric reference frame. The method supports both single- and multi-camera systems without relying on depth sensors or calibration patterns.

Experiments on robotic-arm datasets show that Calib3R provides accurate and robust calibration together with metrically consistent scene reconstructions, even when only a small number of images is available. The formulation is also agnostic to the underlying 3D foundation model and can directly benefit from future advances in this area, as illustrated by the improved results obtained with VGGT.

Future work will explore online and incremental extensions that continuously update the calibration and reconstruction as new visual observations and robot poses become available. Further directions include extending Calib3R to more complex robotic platforms, such as mobile manipulators and multi-robot systems.

\bibliographystyle{IEEEtran}
\bibliography{references}

@INPROCEEDINGS{10186703,
  author={Horn, Markus and Wodtko, Thomas and Buchholz, Michael and Dietmayer, Klaus},
  booktitle={2023 IEEE Intelligent Vehicles Symposium (IV)}, 
  title={Extrinsic Infrastructure Calibration Using the Hand-Eye Robot-World Formulation}, 
  year={2023},
  volume={},
  number={},
  pages={1-8},
  doi={10.1109/IV55152.2023.10186703}}

@article{andreff2001robot,
  title={Robot hand-eye calibration using structure-from-motion},
  author={Andreff, Nicolas and Horaud, Radu and Espiau, Bernard},
  journal={The International Journal of Robotics Research},
  volume={20},
  number={3},
  pages={228--248},
  year={2001},
  publisher={SAGE Publications}
}

@INPROCEEDINGS{5509880,
  author={Gao, Chao and Spletzer, John R.},
  booktitle={2010 IEEE International Conference on Robotics and Automation}, 
  title={On-line calibration of multiple LIDARs on a mobile vehicle platform}, 
  year={2010},
  volume={},
  number={},
  pages={279-284},
  doi={10.1109/ROBOT.2010.5509880}}

@INPROCEEDINGS{5509954,
  author={Antonelli, Gianluca and Caccavale, Fabrizio and Grossi, Flavio and Marino, Alessandro},
  booktitle={2010 IEEE International Conference on Robotics and Automation}, 
  title={Simultaneous calibration of odometry and camera for a differential drive mobile robot}, 
  year={2010},
  volume={},
  number={},
  pages={5417-5422},
  doi={10.1109/ROBOT.2010.5509954}}

@ARTICLE{9390394,
  author={Pedrosa, Eurico and Oliveira, Miguel and Lau, Nuno and Santos, Vítor},
  journal={IEEE Transactions on Robotics}, 
  title={A General Approach to Hand–Eye Calibration Through the Optimization of Atomic Transformations}, 
  year={2021},
  volume={37},
  number={5},
  pages={1619-1633},
  doi={10.1109/TRO.2021.3062306}}

@article{li2024automatic,
  title={Automatic Robot Hand-Eye Calibration Enabled by Learning-Based 3D Vision},
  author={Li, Leihui and Yang, Xingyu and Wang, Riwei and Zhang, Xuping},
  journal={Journal of Intelligent \& Robotic Systems},
  volume={110},
  number={3},
  pages={130},
  year={2024},
  publisher={Springer}
}

@article{rameau2022mc,
  title={MC-Calib: A generic and robust calibration toolbox for multi-camera systems},
  author={Rameau, Francois and Park, Jinsun and Bailo, Oleksandr and Kweon, In So},
  journal={Computer Vision and Image Understanding},
  volume={217},
  pages={103353},
  year={2022},
  publisher={Elsevier}
}

@article{yan2022opencalib,
  title={Opencalib: A multi-sensor calibration toolbox for autonomous driving},
  author={Yan, Guohang and Liu, Zhuochun and Wang, Chengjie and Shi, Chunlei and Wei, Pengjin and Cai, Xinyu and Ma, Tao and Liu, Zhizheng and Zhong, Zebin and Liu, Yuqian and others},
  journal={Software Impacts},
  volume={14},
  pages={100393},
  year={2022},
  publisher={Elsevier}
}

@inproceedings{furgale2013unified,
  title={Unified temporal and spatial calibration for multi-sensor systems},
  author={Furgale, Paul and Rehder, Joern and Siegwart, Roland},
  booktitle={2013 IEEE/RSJ International Conference on Intelligent Robots and Systems},
  pages={1280--1286},
  year={2013},
  organization={IEEE}
}

@inproceedings{wang2024dust3r,
  title={Dust3r: Geometric 3d vision made easy},
  author={Wang, Shuzhe and Leroy, Vincent and Cabon, Yohann and Chidlovskii, Boris and Revaud, Jerome},
  booktitle={Proceedings of the IEEE/CVF Conference on Computer Vision and Pattern Recognition},
  pages={20697--20709},
  year={2024}
}

@inproceedings{leroy2024grounding,
  title={Grounding image matching in 3d with mast3r},
  author={Leroy, Vincent and Cabon, Yohann and Revaud, J{\'e}r{\^o}me},
  booktitle={European Conference on Computer Vision},
  pages={71--91},
  year={2024},
  organization={Springer}
}

@ARTICLE{10874177,
  author={Liu, Yangjun and Liu, Sheng and Chen, Binghan and Yang, Zhi-Xin and Xu, Sheng},
  journal={IEEE Transactions on Robotics}, 
  title={Fusion-Perception-to-Action Transformer: Enhancing Robotic Manipulation With 3-D Visual Fusion Attention and Proprioception}, 
  year={2025},
  volume={41},
  number={},
  pages={1553-1567},
  doi={10.1109/TRO.2025.3539193}}

@inproceedings{barcellona2025dream,
  title={Dream to manipulate: Compositional world models empowering robot imitation learning with imagination},
  author={Barcellona, Leonardo and Zadaianchuk, Andrii and Allegro, Davide and Papa, Samuele and Ghidoni, Stefano and Gavves, Efstratios},
  booktitle={International Conference on Learning Representations},
  volume={2025},
  pages={56729--56763},
  year={2025}
}

@ARTICLE{10418577,
  author={Huang, Yixuan and Taylor, Nichols Crawford and Conkey, Adam and Liu, Weiyu and Hermans, Tucker},
  journal={IEEE Transactions on Robotics}, 
  title={Latent Space Planning for Multiobject Manipulation With Environment-Aware Relational Classifiers}, 
  year={2024},
  volume={40},
  number={},
  pages={1724-1739},
  doi={10.1109/TRO.2024.3360956}}

@article{allegro2024multi,
  title={Multi-camera hand-eye calibration for human-robot collaboration in industrial robotic workcells},
  author={Allegro, Davide and Terreran, Matteo and Ghidoni, Stefano},
  journal={IEEE Robotics and Automation Letters},
  year={2024},
  publisher={IEEE}
}

@article{tsai1989new,
  title={A new technique for fully autonomous and efficient 3 d robotics hand/eye calibration},
  author={Tsai, Roger Y and Lenz, Reimar K and others},
  journal={IEEE Transactions on robotics and automation},
  volume={5},
  number={3},
  pages={345--358},
  year={1989}
}

@article{park1994robot,
  title={Robot sensor calibration: solving AX= XB on the Euclidean group},
  author={Park, Frank C and Martin, Bryan J},
  journal={IEEE Transactions on Robotics and Automation},
  volume={10},
  number={5},
  pages={717--721},
  year={1994},
  publisher={IEEE}
}

@article{horaud1995hand,
  title={Hand-eye calibration},
  author={Horaud, Radu and Dornaika, Fadi},
  journal={The international journal of robotics research},
  volume={14},
  number={3},
  pages={195--210},
  year={1995},
  publisher={Sage Publications Sage CA: Thousand Oaks, CA}
}

@inproceedings{valassakis2022learning,
  title={Learning eye-in-hand camera calibration from a single image},
  author={Valassakis, Eugene and Dreczkowski, Kamil and Johns, Edward},
  booktitle={Conference on Robot Learning},
  pages={1336--1346},
  year={2022},
  organization={PMLR}
}

@article{shah2013solving,
  title={Solving the robot-world/hand-eye calibration problem using the Kronecker product},
  author={Shah, Mili},
  journal={Journal of Mechanisms and Robotics},
  volume={5},
  number={3},
  pages={031007},
  year={2013},
  publisher={American Society of Mechanical Engineers}
}

@inproceedings{andreff1999line,
  title={On-line hand-eye calibration},
  author={Andreff, Nicolas and Horaud, Radu and Espiau, Bernard},
  booktitle={Second International Conference on 3-D Digital Imaging and Modeling (Cat. No. PR00062)},
  pages={430--436},
  year={1999},
  organization={IEEE}
}

@inproceedings{daniilidis1996dual,
  title={The dual quaternion approach to hand-eye calibration},
  author={Daniilidis, Konstantinos and Bayro-Corrochano, Eduardo},
  booktitle={Proceedings of 13th International Conference on Pattern Recognition},
  volume={1},
  pages={318--322},
  year={1996},
  organization={IEEE}
}

@article{li2010simultaneous,
  title={Simultaneous robot-world and hand-eye calibration using dual-quaternions and Kronecker product},
  author={Li, Aiguo and Wang, Lin and Wu, Defeng},
  journal={Int. J. Phys. Sci},
  volume={5},
  number={10},
  pages={1530--1536},
  year={2010}
}

@inproceedings{evangelista2022unified,
  title={An unified iterative hand-eye calibration method for eye-on-base and eye-in-hand setups},
  author={Evangelista, Daniele and Allegro, Davide and Terreran, Matteo and Pretto, Alberto and Ghidoni, Stefano},
  booktitle={2022 IEEE 27th International Conference on Emerging Technologies and Factory Automation (ETFA)},
  pages={1--7},
  year={2022},
  organization={IEEE}
}

@inproceedings{allegro2024memroc,
  title={MEMROC: Multi-Eye to Mobile RObot Calibration},
  author={Allegro, Davide and Terreran, Matteo and Ghidoni, Stefano},
  booktitle={2024 IEEE/RSJ International Conference on Intelligent Robots and Systems (IROS)},
  pages={884--891},
  year={2024},
  organization={IEEE}
}

@article{duisterhof2024mast3r,
  title={MASt3R-SfM: a Fully-Integrated Solution for Unconstrained Structure-from-Motion},
  author={Duisterhof, Bardienus and Zust, Lojze and Weinzaepfel, Philippe and Leroy, Vincent and Cabon, Yohann and Revaud, Jerome},
  journal={arXiv preprint arXiv:2409.19152},
  year={2024}
}

@article{zhi2024unifying,
  title={Unifying representation and calibration with 3d foundation models},
  author={Zhi, Weiming and Tang, Haozhan and Zhang, Tianyi and Johnson-Roberson, Matthew},
  journal={IEEE Robotics and Automation Letters},
  volume={9},
  number={10},
  pages={8953--8960},
  year={2024},
  publisher={IEEE}
}

@inproceedings{wang2022accurate,
  title={Accurate calibration of multi-perspective cameras from a generalization of the hand-eye constraint},
  author={Wang, Yifu and Jiang, Wenqing and Huang, Kun and Schwertfeger, S{\"o}ren and Kneip, Laurent},
  booktitle={2022 International Conference on Robotics and Automation (ICRA)},
  pages={1244--1250},
  year={2022},
  organization={IEEE}
}

@inproceedings{schonberger2016structure,
  title={Structure-from-motion revisited},
  author={Schonberger, Johannes L and Frahm, Jan-Michael},
  booktitle={Proceedings of the IEEE conference on computer vision and pattern recognition},
  pages={4104--4113},
  year={2016}
}

@inproceedings{fang2020graspnet,
  title={GraspNet-1Billion: A Large-Scale Benchmark for General Object Grasping},
  author={Fang, Hao-Shu and Wang, Chenxi and Gou, Minghao and Lu, Cewu},
  booktitle={Proceedings of the IEEE/CVF Conference on Computer Vision and Pattern Recognition},
  pages={11444--11453},
  year={2020}
}

@inproceedings{zhang2018deep,
  title={Deep depth completion of a single rgb-d image},
  author={Zhang, Yinda and Funkhouser, Thomas},
  booktitle={Proceedings of the IEEE conference on computer vision and pattern recognition},
  pages={175--185},
  year={2018}
}

@article{min2024multi,
  title={Multi-camera unified pre-training via 3d scene reconstruction},
  author={Min, Chen and Xiao, Liang and Zhao, Dawei and Nie, Yiming and Dai, Bin},
  journal={IEEE Robotics and Automation Letters},
  volume={9},
  number={4},
  pages={3243--3250},
  year={2024},
  publisher={IEEE}
}

@inproceedings{zhu2020autonomous,
  title={Autonomous robot navigation based on multi-camera perception},
  author={Zhu, Kunyan and Chen, Wei and Zhang, Wei and Song, Ran and Li, Yibin},
  booktitle={2020 IEEE/RSJ International Conference on Intelligent Robots and Systems (IROS)},
  pages={5879--5885},
  year={2020},
  organization={IEEE}
}

@inproceedings{heller2011structure,
  title={Structure-from-motion based hand-eye calibration using {$L_\infty$} minimization},
  author={Heller, Jan and Havlena, Michal and Sugimoto, Akihiro and Pajdla, Tomas},
  booktitle={CVPR 2011},
  pages={3497--3503},
  year={2011},
  organization={IEEE}
}

@article{furukawa2015multi,
  title={Multi-view stereo: A tutorial},
  author={Furukawa, Yasutaka and Hern{\'a}ndez, Carlos and others},
  journal={Foundations and trends{\textregistered} in Computer Graphics and Vision},
  volume={9},
  number={1-2},
  pages={1--148},
  year={2015},
  publisher={Now Publishers, Inc.}
}

@inproceedings{wang2024vggsfm,
  title={Vggsfm: Visual geometry grounded deep structure from motion},
  author={Wang, Jianyuan and Karaev, Nikita and Rupprecht, Christian and Novotny, David},
  booktitle={Proceedings of the IEEE/CVF conference on computer vision and pattern recognition},
  pages={21686--21697},
  year={2024}
}

@inproceedings{he2024detector,
  title={Detector-free structure from motion},
  author={He, Xingyi and Sun, Jiaming and Wang, Yifan and Peng, Sida and Huang, Qixing and Bao, Hujun and Zhou, Xiaowei},
  booktitle={Proceedings of the IEEE/CVF Conference on Computer Vision and Pattern Recognition},
  pages={21594--21603},
  year={2024}
}

@article{chen2025vidbot,
  title={VidBot: Learning Generalizable 3D Actions from In-the-Wild 2D Human Videos for Zero-Shot Robotic Manipulation},
  author={Chen, Hanzhi and Sun, Boyang and Zhang, Anran and Pollefeys, Marc and Leutenegger, Stefan},
  journal={arXiv preprint arXiv:2503.07135},
  year={2025}
}

@inproceedings{evangelista2023graph,
  title={A graph-based optimization framework for hand-eye calibration for multi-camera setups},
  author={Evangelista, Daniele and Olivastri, Emilio and Allegro, Davide and Menegatti, Emanuele and Pretto, Alberto},
  booktitle={2023 IEEE International Conference on Robotics and Automation (ICRA)},
  pages={11474--11480},
  year={2023},
  organization={IEEE}
}

@inproceedings{wang2025vggt,
  title={Vggt: Visual geometry grounded transformer},
  author={Wang, Jianyuan and Chen, Minghao and Karaev, Nikita and Vedaldi, Andrea and Rupprecht, Christian and Novotny, David},
  booktitle={Proceedings of the Computer Vision and Pattern Recognition Conference},
  pages={5294--5306},
  year={2025}
}

@article{bauer2024challenges,
  title={Challenges for monocular 6d object pose estimation in robotics},
  author={Bauer, Dominik and H{\"o}nig, Peter and Weibel, Jean-Baptiste and Garc{\'\i}a-Rodr{\'\i}guez, Jos{\'e} and Vincze, Markus and others},
  journal={IEEE Transactions on Robotics},
  year={2024},
  publisher={IEEE}
}

@inproceedings{an2024rgbmanip,
  title={Rgbmanip: Monocular image-based robotic manipulation through active object pose estimation},
  author={An, Boshi and Geng, Yiran and Chen, Kai and Li, Xiaoqi and Dou, Qi and Dong, Hao},
  booktitle={2024 IEEE International Conference on Robotics and Automation (ICRA)},
  pages={7748--7755},
  year={2024},
  organization={IEEE}
}

\end{document}